\documentclass{article} 
\usepackage[preprint]{colm2025_conference}

\usepackage{microtype}
\usepackage{hyperref}
\usepackage{url}
\usepackage{booktabs}
\usepackage{amsmath}
\usepackage{mathtools}
\usepackage{graphicx}

\usepackage{lineno}
\usepackage{wrapfig}
\usepackage{mdframed}
\usepackage{fvextra}    
\usepackage{fancyvrb}      
\usepackage{placeins}

\definecolor{darkblue}{rgb}{0, 0, 0.5}
\hypersetup{colorlinks=true, citecolor=darkblue, linkcolor=darkblue, urlcolor=darkblue}
\usepackage[most]{tcolorbox}

\title{Uncalibrated Reasoning: GRPO Induces Overconfidence for Stochastic Outcomes}

\author{Michael Bereket \& Jure Leskovec\\
Department of Computer Science\\
Stanford University\\
\texttt{\{mbereket, jure\}@cs.stanford.edu}
}

\begin{document}

\ifcolmsubmission
\linenumbers
\fi

\maketitle

\begingroup
\renewcommand\thefootnote{}%
\footnotetext{Code available at \url{https://github.com/mbereket/uncalibrated_reasoning}}%
\addtocounter{footnote}{-1}%
\endgroup

\begin{abstract}
Reinforcement learning (RL) has proven remarkably effective at improving the accuracy of language models in verifiable and deterministic domains like mathematics. Here, we examine if current RL methods are also effective at optimizing language models in verifiable domains with stochastic outcomes, like scientific experiments. Through applications to synthetic data and real-world biological experiments, we demonstrate that Group Relative Policy Optimization (GRPO) induces overconfident probability predictions for binary stochastic outcomes, while Proximal Policy Optimization (PPO) and REINFORCE Leave-One-Out (RLOO) yield well-calibrated models. We show that removing group standard normalization in GRPO fixes its miscalibration and provide a theoretical explanation for why normalization causes overconfidence. Our results provide new evidence against the use of standard normalization in GRPO and help pave the way for applications of RL for reasoning language models beyond deterministic domains.
\end{abstract}

\section{Introduction}

Reinforcement learning (RL) has achieved remarkable success at improving the accuracy of language models in verifiable domains like mathematics and coding \citep{o1, deepseekmath, kimi}. In particular, recent success has been achieved by optimizing language models to generate chain-of-thought text before responding to a prompt (often called "reasoning") with supervision from a verifier. Current research has focused primarily on domains where proposed answers are deterministically correct or incorrect.

We posit that an important next step for the reasoning RL paradigm is to expand to domains with verifiable yet stochastic answers. For example, scientific experiments, which are subject to random variation, could serve as powerful verifiers for optimizing language models beyond current written knowledge. Scientific reasoning models trained in this manner could support hypothesis generation, experimental design, and decision making through both their predictions and generated reasoning traces. Other potentially impactful settings for training reasoning models with stochastic outcomes include model alignment, which considers human behaviors and preferences \citep{rlhf, instructgpt}, and even model uncertainty estimation, which is important for high-stakes decision making and can be framed as modeling the probability that a prediction is correct \citep{bandLinguisticCalibrationLongForm2024, stangelRewardingDoubtReinforcement2025, rlcr2025}.

\begin{figure}
    \centering
    \includegraphics[width=0.95\linewidth]{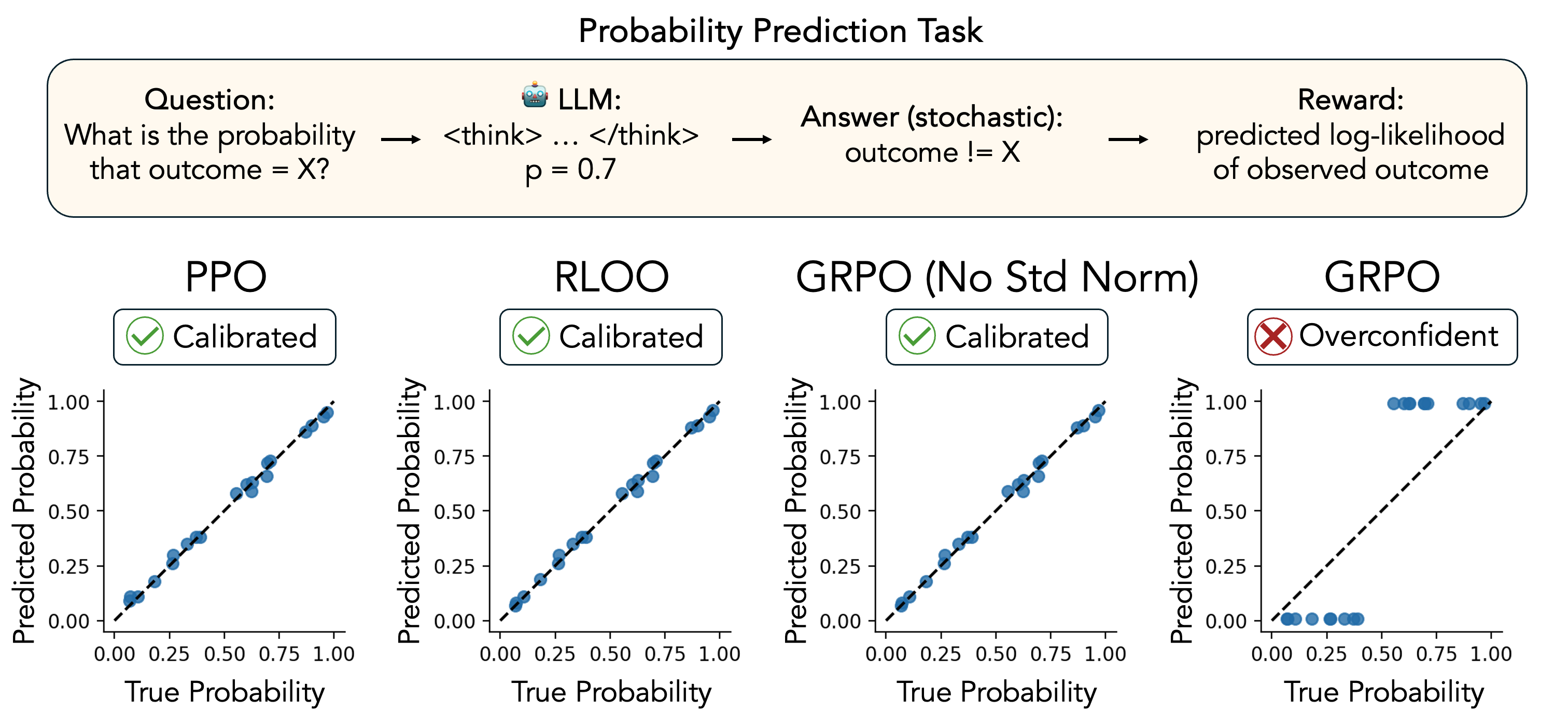}
    \caption{Group standard normalization in GRPO induces overconfident predictions of stochastic outcome probabilities. \textbf{Top:} Probability prediction task. \textbf{Bottom:} Synthetic data experiment results. Models trained with PPO, RLOO, and GRPO with no standard normalization are well calibrated, while models trained with GRPO are extremely overconfident.}
    \label{fig:key_result}
\end{figure}

In this paper, we examine whether three popular algorithms for reasoning RL in deterministic domains, namely GRPO \citep{deepseekmath}, PPO \citep{ppo}, and RLOO \citep{rloo1, rloo2}, are also effective in settings with binary stochastic outcomes. Through applications to synthetic data and real-world biological experiments, we demonstrate that models trained to maximize observation likelihoods with GRPO predict highly overconfident outcome probabilities, while models optimized with PPO and RLOO are well calibrated (Fig.~\ref{fig:key_result} and \ref{fig:pert_preds}). We find that GRPO can be modified for better calibration by removing the group standard normalization term and provide a theoretical justification for why normalization causes overconfidence. In sum, our results provide new evidence against the use of standard normalization in GRPO, highlight the value of unbiasedness as a design principle for policy gradients, and help support future applications of reasoning RL beyond deterministic domains.

\section{Preliminaries} \label{Preliminaries}


\textbf{RL with Language Models} Reinforcement learning methods cast autoregressive language models as stochastic policies $\pi_\theta$ that specify actions (selecting new tokens) based on the current state (the prompt and prior generated tokens). We consider a setting with outcome supervision, where the goal is to maximize the expected reward received from a verifier that scores the correctness of a response given the ground-truth answer. While current work focuses primarily on settings with deterministic answers, we consider answers that may be stochastic conditional on the prompt.

\textbf{Value and Advantage Functions} The state value function $V^{\pi}(s)$ is defined as the expected reward from following policy $\pi$ from state $s$, and the state-action value function $Q^{\pi}(s,a)$ is defined as the expected reward of following policy $\pi$ from state $s$ when the next action is set to be $a$. The advantage function $A^{\pi}(s, a) = Q^\pi(s, a) - V^\pi(s)$ is the expected increase in reward from selecting $a$ as the next action from state $s$ relative to an action sampled from $\pi$.

\textbf{Policy Gradients} Policy gradient methods optimize policy $\pi_\theta$ by directly estimating the gradient of the expected reward with respect to policy parameters. Let $q$ be a prompt, $a$ be the true answer, $\mathbf{o}=(o_1, ..., o_t)$ be a sequence of response tokens, and $\mathrm{r}(\mathbf{o},a)$ be the final reward received from the verifier. From the policy gradient theorem \citep{suttonPolicyGradientMethods1999}

\begin{equation}
    \hat{g}^{\text{PG}} = \hat{\mathbb{E}}_{\substack{q\sim p(Q),\ a\sim{p(A|q)},\\\mathbf{o}\sim \pi_\theta(O|q)}} \left[ \sum_{t=1}^{|\mathbf{o}|} \nabla_\theta \log \pi_\theta(o_t | s_t) \left(\mathrm{r}(\mathbf{o},a) - b(s_t)\right) \right]
  \label{eq:vanilla-pg}
\end{equation}

is an unbiased estimate of the policy gradient, where $\hat{\mathbb{E}}$ is an empirical sample mean,  $s_t \coloneqq (q,o_{<t})$ is the state at step $t$ (the prompt and prior tokens), and baseline $b(s_t)$ is a function of the current state. A common choice is $b(s_t)=\hat{V}(s_t)$, which makes the baselined reward equivalent to an estimate of the advantage $\hat{A}(s_t,o_t)$. The policy gradient estimator can be interpreted as as increasing the probability of actions with above average expected rewards and decreasing the probability of actions with below average expected rewards.

Each of the three algorithms considered in this paper (GRPO, PPO, and RLOO) are policy gradient methods. We discuss the different strategies these methods take for advantage estimation and deviations from the policy gradient estimator $\hat{g}^{\text{PG}}$ below.

\textbf{Advantage Estimation for Policy Gradients} Consider sampling $G$ responses from a single prompt, and let $\mathbf{r}=(r_1, ..., r_G)$ be the rewards for these responses. Let $\hat{A}_{i,t}$ be the estimated advantage for token $t$ in response $i$. PPO, RLOO, and GRPO then have the following advantage estimators:

\begin{table}[h]
  \centering
  {
    \normalsize
    \renewcommand{\arraystretch}{1.2}  
    \begin{tabular}{@{}l c c@{}}
      Algorithm & Advantage estimator $\hat{A}_{i,t}$ & Unbiased PG?\\
      \midrule
      PPO & $r_i - \hat{V}_\psi(s_{i,t})$ & Yes\\
      RLOO & $r_i - \operatorname{mean}(\mathbf{r}_{j \neq i})$ & Yes\\
      GRPO & $\displaystyle\tfrac{r_i - \operatorname{mean}(\mathbf{r})}{\operatorname{std}(\mathbf{r}) + \epsilon}$ & No\\
        GRPO (No Std Norm) & $r_i - \operatorname{mean}(\mathbf{r})$ & No (proportional)\\
    \end{tabular}
      }
\end{table}

PPO uses Generalized Advantage Estimation (GAE) \citep{gae} and learns an explicit model of the value function $\hat{V}_\psi$ as a baseline (we focus on the unbiased variant of GAE). To avoid the computational costs associated with learning an explicit value model, RLOO and GRPO instead compute a Monte Carlo estimate of the value using multiple responses generated from the same prompt. Specifically, RLOO subtracts the mean reward from the other sampled responses, yielding an unbiased advantage estimate, while GRPO subtracts the mean reward from all responses and divides by the standard deviation, which is biased. We also consider a variant of GRPO without standard normalization, which yields a policy gradient estimate that is proportional to an unbiased estimate. We note that RLOO and GRPO uses the same advantage estimate for each token, which can be interpreted as casting question answering as a bandit problem where generating the full response corresponds to a single action.

\textbf{Clipped Policy Gradients} The primary contribution of PPO was to introduce a clipped policy gradient estimator to stabilize training when performing multiple gradient updates on a single batch of rollouts (at the cost of introducing bias). The clipped estimator is

\begin{align*}
    \hat{g}_t^{\text{clip}} = \nabla_\theta \hat{\mathbb{E}}_{\substack{q\sim p(Q)\\\mathbf{o}\sim \pi_{\theta_{old}}(O|q)}} \text{min} \left[ \frac{\pi_\theta(o_{t}|q,o_{<t})}{\pi_{\theta_{old}}(o_{t}|q,o_{<t})} \hat{A}_t, \text{clip} \left( \frac{\pi_\theta(o_{t}|q,o_{<t})}{\pi_{\theta_{old}}(o_{t}|q,o_{<t})}, 1-\epsilon, 1+\epsilon  \right) \hat{A}_{t} \right]
\end{align*}

When applied on-policy, $\pi_\theta = \pi_{\theta_{old}}$ and the clipped estimator reduces to the vanilla policy gradient. The clipped policy gradient is also used in GRPO and can be applied with any of the advantage estimators discussed above.

\section{Experiments}

\subsection{Problem Statement}

We consider the following \textbf{probability prediction task}: given a prompt $q$ and binary answer $a \in \{0, 1\}$, predict the probability that $a=1$. Training data consists of question-answer pairs $(q_i, a_i)|_{i=1}^N$, where observed answers $a_i$ are sampled from some unknown probability distribution $a_i \sim p(A|q_i)$. We apply RL for this task with the reward function defined as the log-likelihood of the observed answer under the model predicted probability.

\subsection{Metrics}

We evaluate model predictions for both calibration and classification performance. To measure calibration, we visualize reliability plots and compute the Expected Calibration Error (ECE). ECE is computed by binning predicted probabilities (we use 10 bins) and computing the average difference between the frequency of positive instances and mean predicted probability in each bin, weighted by the number of points. We measure classification performance with both the Area Under the Receiver Operator Characteristic (AUROC) and accuracy of predicted probabilities thresholded at 0.5.

\subsection{Experiment 1: Synthetic Data} \label{Toy}

\begin{table}[t]
  \centering
  \setlength{\tabcolsep}{6pt}       
  \renewcommand{\arraystretch}{1.15} 
  \begin{tabular}{l *{3}{c} *{3}{c}}
    \toprule
    & \multicolumn{3}{c}{\textbf{Synthetic Data}} 
    & \multicolumn{3}{c}{\textbf{CRISPR Screen}} \\
    \cmidrule(lr){2-4}\cmidrule(lr){5-7}
    \textbf{Algorithm} & ECE ($\downarrow$) & AUROC ($\uparrow$) & Acc. ($\uparrow$) & ECE ($\downarrow$) & AUROC ($\uparrow$) & Acc. ($\uparrow$) \\
    \midrule
    GRPO & 0.239 & 0.75 & 0.75 & 0.292 & 0.69 & 0.67 \\
    GRPO (No Std.) & 0.002 & 0.82 & 0.75 & 0.036 & 0.72 & 0.68 \\
    RLOO & 0.002 & 0.82 & 0.75 & 0.040 & 0.72 & 0.68 \\
    PPO & 0.005 & 0.82 & 0.75 & 0.038 & 0.72 & 0.67 \\
    \bottomrule
  \end{tabular}
  \caption{Evaluation metrics from probability prediction experiments. Across applications to synthetic data and real-world biological experiments, we find that GRPO achieves poor ECE and AUROC relative to GRPO without standard normalization, RLOO, and PPO. All algorithms perform nearly identically on accuracy with predicted probabilities thresholded at 0.5, which does not require well-calibrated predictions.}
  \label{tab:results}
\end{table}

We begin by characterizing the behavior of each RL algorithm in a minimal synthetic data experiment with known ground-truth probabilities. 

\textbf{Data} We simulate a dataset of 10,000 $(q_i, c_i, a_i)$ triples, representing questions, categories, and binary answers. Questions are randomly assigned to one of 20 random categories. For each category, a true category answer rate is sampled from a uniform distribution: $p_1, ..., p_{20} \sim \text{Uniform}(0, 1)$. Answers are then sampled from the true answer rate for the question category: $a_i | q_i, c_i \sim \text{Bernoulli}(p_{c_i})$.

\textbf{Model} We define a minimal "language model" that samples a single token representing the predicted probability given a question. We parameterize the model as a categorical distribution $p_\theta(a_i=1|q_i) = p_\theta(a_i=1|c_i)$, with a learnable parameter for each category / probability token pair. We use a vocabulary of 99 tokens representing probabilities between 0.01 and 0.99. For experiments with PPO, we define a value model that predicts $\hat{V}(q_i) = \psi_{c_i}$, where $\psi_c$ is a learnable parameter for each category.

\textbf{Optimization} We optimize models using PPO, RLOO, GRPO, and GRPO without standard normalization both on-policy and off-policy (1 and 10 gradient updates per rollout, respectively). Off-policy models are optimized with the clipped policy gradient estimator, and we consider clipping thresholds of 0.2 and 0.001 to assess the effects of different clipping rates.

 \textbf{Results} Across all settings, we find that GRPO yields highly overconfident probability predictions: models optimized with GRPO converge to predict the minimum available probability for categories with true probability $<0.5$ and the maximum available probability for categories with true probability $>0.5$ (Fig.~\ref{fig:key_result}). In contrast, GRPO without standard normalization, PPO, and RLOO all yield well-calibrated predictions (Fig.~\ref{fig:key_result}). These observations are reflected GRPO's poor ECE (0.24 vs $<$0.01) and AUROC (0.75 vs 0.82) relative to the other algorithms (Tbl.~\ref{tab:results}). We observe that all considered algorithms perform equivalently on thresolded accuracy, which does not require calibrated predictions. We also obtain nearly identical results when training on-policy and off-policy, even when introducing high clipping rates, which suggests that the clipped policy gradient estimator does not introduce a systematic bias for probability prediction (Appendix Tbl.~\ref{tab:extended_synthetic_results} and Fig.~\ref{fig:synthetic_training_metrics}).

\subsection{Experiment 2: Real-World CRISPR Screen}

\begin{figure}
    \centering
    \includegraphics[width=0.9\linewidth]{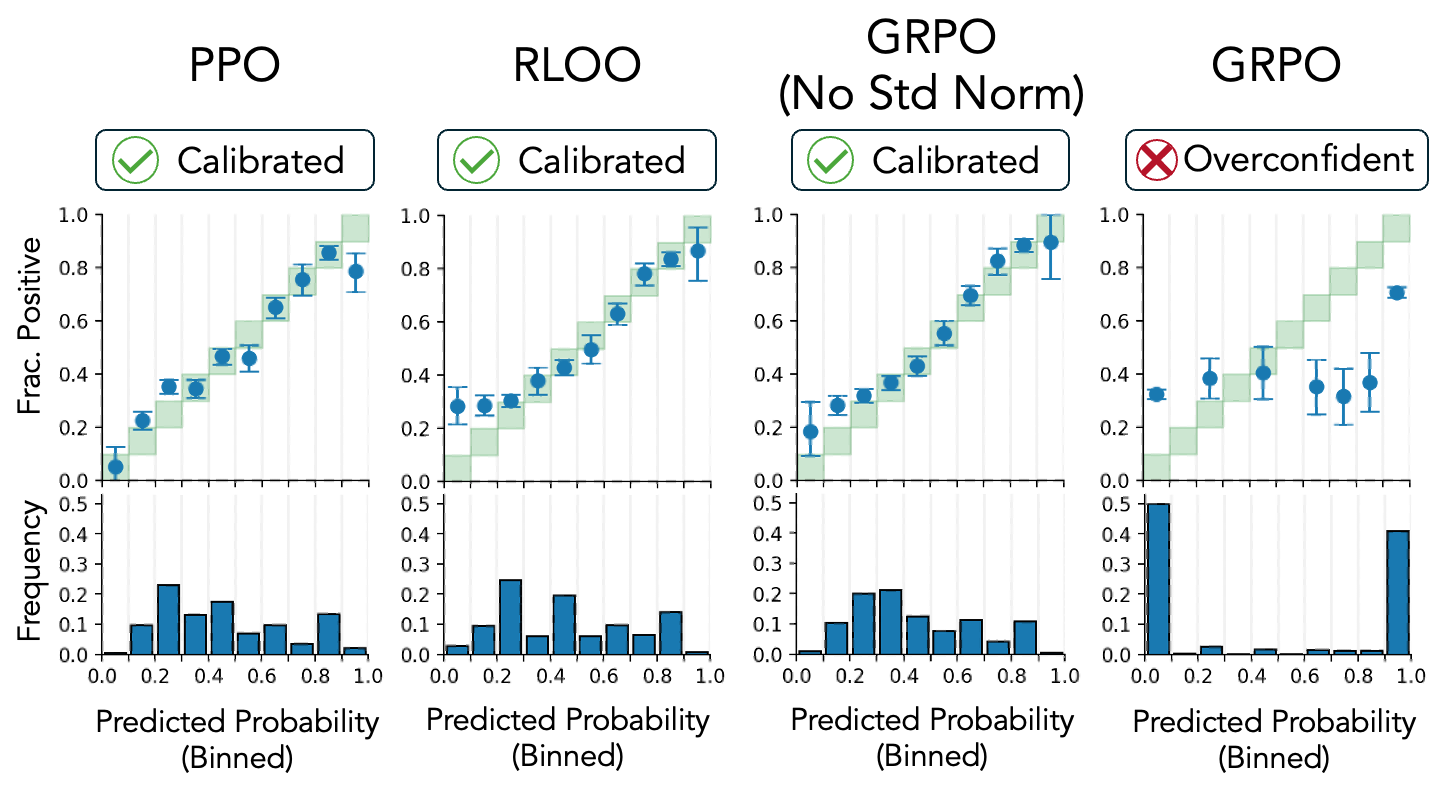}
    \caption{Real-world biological experiment prediction results. We optimize Qwen3-4B to predict the probability of binary experimental outcomes (will perturbing a target gene have a strong effect on a specified phenotype in cells?) with rewards derived from real-world experiments. We find that models optimized with PPO, RLOO, and GRPO with no standard normalization achieve well-calibrated predictions for held-out test perturbations, while GRPO predicts highly overconfident probabilities. Error bars represent 95\% confidence intervals.}
    \label{fig:pert_preds}
\end{figure}

Next, we evaluate if the conclusions from the synthetic data experiment hold when optimizing a large language model (Qwen3-4B, \citet{qwen3}) to predict outcome probabilities in real-world biological experiments.

\textbf{Data} A key challenge in drug discovery is the identification of genetic causal effects: if a gene is perturbed with a drug, how will it affect disease state? In recent years, perturb-seq \citep{dixitPerturbSeqDissectingMolecular2016} has emerged as a powerful experimental technique for identifying causal effects. CRISPR perturb-seq experiments involves perturbing genes with CRISPR and measuring the effect of those perturbations on gene expression counts for each gene in individual cells (which can be interpreted as a broad measurement of cell state). For this experiment, we convert a large perturb-seq dataset from \citet{replogle} into a binary task: for a given perturbed gene and target gene expression phenotype, predict the probability that the perturbed gene has a strong effect on the phenotype (full preprocessing details in Appendix~\ref{crispr_preprocessing}). We sample a balanced dataset of positive and negative instances for the final dataset and generate validation and test splits with held-out perturbations. 

\textbf{Model} We optimize Qwen3-4B to predict the probability that a perturbed gene has a strong effect on a target phenotype. The model is prompted to predict the probability as a percentage between 1 and 99 (full prompt in Appendix~\ref{crispr_prompt}).

\textbf{Optimization} We optimize models with PPO, RLOO, GRPO, and GRPO without standard normalization using Verl \citep{verl}. Each algorithm is applied off-policy (8 updates per sampled training batch) with the clipped policy gradient estimator. Models are trained for 16 epochs with train batch size 512 and 4 rollouts per sample (details in Appendix~\ref{crispr_experiment_details}).

\textbf{Results} Consistent with the synthetic data experiment, we find that optimization with GRPO results in highly overconfident probability predictions (ECE=0.29), while GRPO with no standard normalization, PPO, and RLOO yield well-calibrated models (ECE$\leq$0.04, Fig.~\ref{fig:pert_preds} and Tbl.~\ref{tab:results}). GRPO again performs poorly on AUROC (0.69 vs 0.72 from the other algorithms) and all models are similarly accurate. We also find that the clipped policy gradient, which was used for all models in this experiment, did not cause biased probability predictions.

\section{Theoretical Analysis}
\label{sec:theoretical_analysis}

\begin{figure}
    \centering
    \includegraphics[width=0.95\linewidth]{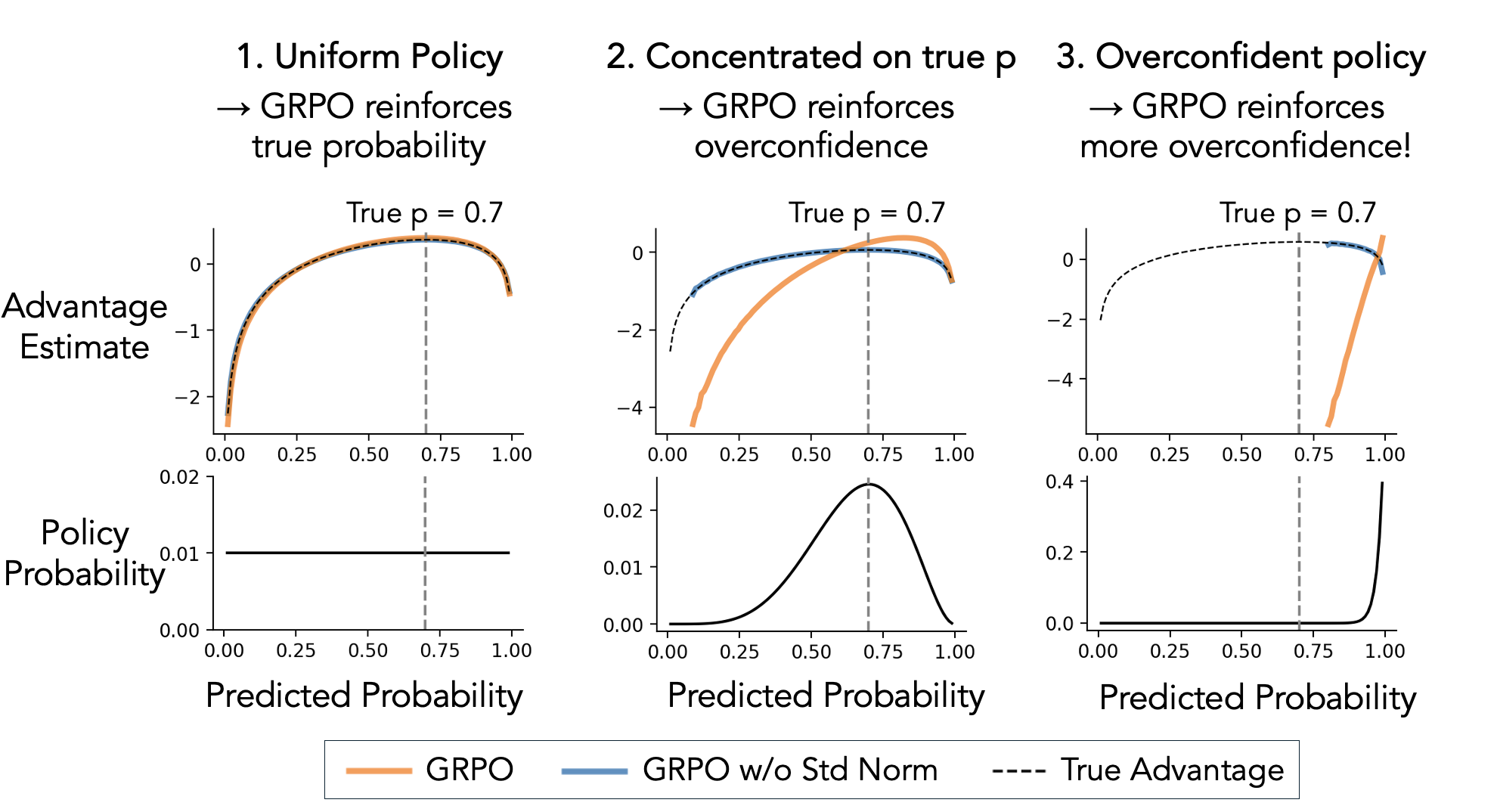}
    \caption{Bias in GRPO advantage estimates explains overconfident predictions. Advantages are computed with a log-likelihood reward.
    \textbf{Left:} Under a uniform policy, both GRPO and GRPO without standard normalization closely approximate the true advantages. \textbf{Middle:} Under a policy concentrated on the true probability, GRPO overestimates the advantage of overconfident predictions. \textbf{Right:} As the policy becomes increasingly overconfident, GRPO increasingly overestimates the advantage of more overconfident predictions.  This pattern creates a positive feedback loop towards increasingly overconfident predictions consistent with our experimental observations. 
    }
    \label{fig:cause}
\end{figure}

Finally, we analyze why standard normalization in GRPO induces overconfident predictions. Recall that GRPO reinforces actions based on their estimated advantage: actions that have large advantages are made more likely, while actions with negative advantages are made less likely. We will show that standard normalization causes GRPO to overestimate the advantage of overconfident predictions, resulting in overconfident policies (Fig.~\ref{fig:cause}).

In Appendix \ref{bias_analysis}, we derive expressions for the expected advantage estimates from GRPO with and without standard normalization. Let $q$ be a prompt with stochastic answers $a \sim \text{Bernoulli}(p)$, let $\hat{p}$ be a predicted probability that $a=1$, and let $\mathrm{r}(\hat{p}, a)$ be a reward function such as the log-likelihood reward $\mathrm{r}(\hat{p}, a)=a\log\hat{p} + (1-a)\log(1-\hat{p})$. The true advantage for prediction $\hat{p}$ is then
\begin{align*}
A(q,\hat{p}) = p(\mathrm{r}(\hat{p}, 1)) - \mu_1) + (1-p)(\mathrm{r}(\hat{p}, 0) - \mu_0)
\end{align*}
where $\mu_1 = \mathbb{E}_{\hat{p}'\sim \pi_\theta(q)}\left[\mathrm{r}(\hat{p}', 1)\right]$ and $\mu_0 = \mathbb{E}_{\hat{p}'\sim \pi_\theta(q)}\left[\mathrm{r}(\hat{p}', 0)\right]$. We show that the expected advantage estimate for GRPO without standard normalization is
\begin{align*}
\mathbb{E} \left[ \hat{A}^{\text{NO-STD}}(q,\hat{p})\right] = \frac{G-1}{G} A(q,\hat{p}) \propto A(q,\hat{p})
\end{align*}
This means that the policy gradients using GRPO without standard normalization are approximately unbiased (up to a constant factor), consistent with the calibrated predictions we observed experimentally. In contrast, the advantage estimate for GRPO is approximately

\begin{align*}
\mathbb{E}\left[\hat{A}^{\text{STD}}(q,\hat{p})\right] \approx \frac{1}{\sigma_1 + \epsilon}p(\mathrm{r}(\hat{p}, 1)) - \mu_1) + \frac{1}{\sigma_0 + \epsilon}(1-p)(\mathrm{r}(\hat{p}, 0) - \mu_0)
\end{align*}

where $\sigma_1 = \mathbb{E}_{\hat{\mathbf{p}}}\left[\text{std}(\mathrm{r}(\hat{\mathbf{p}}, 1))\right]$, $\sigma_0 = \mathbb{E}_{\hat{\mathbf{p}}}\left[\text{std}(\mathrm{r}(\hat{\mathbf{p}}, 0))\right]$ and $\hat{\mathbf{p}}$ denotes samples of $G$ model predictions from  the prompt. We observe that the approximate GRPO advantage expression closely resembles the true advantage with the addition of $\frac{1}{\sigma_0+\epsilon}$ and $\frac{1}{\sigma_1+\epsilon}$ coefficients, which introduce a policy-dependent bias that we analyze empirically.

In Fig.~\ref{fig:cause}, we visualize empirical estimates of the expected advantage for GRPO with and without standard normalization with a log-likelihood reward (estimation details in \ref{empirical_advantage_details}).  Under a uniform policy, the advantage estimates from both methods closely approximate the true advantage (left column). As the policy begins to concentrate around the true probability, we observe that GRPO starts to overestimate the advantage of overconfident predictions, while the unnormalized estimates remains accurate (center column). This will cause GRPO to reinforce overconfident predictions more strongly than the true probability, resulting in overconfident policies. Finally, we observe that under a very overconfident policy, GRPO's advantage estimates will have an even more extreme bias towards overconfident predictions, while GRPO without standard normalization remains approximately unbiased (right column). These observations are consistent with our approximate GRPO advantage expression: as the policy concentrates above 0.5, $\sigma_0$ becomes larger than $\sigma_1$, resulting in a reduced weight on the penalty for overconfident predictions (Appendix Fig.~\ref{fig:explanation_sigma}).   

To summarize, group standard normalization in GRPO's advantage estimates creates a policy-dependent bias that pushes policies towards overconfident predictions. While our analysis focused on a log-likelihood reward, we also consider rewards based on other strictly proper scoring rules in Appendix~\ref{other_reward_bias}.

\section{Discussion}

Many important tasks, from scientific experimentation to preference modeling, require reasoning about the likelihood of stochastic outcomes. We showed that reasoning language models optimized to predict the probability of binary stochastic outcomes from samples with GRPO are highly overconfident, while models optimized with PPO and RLOO are well calibrated (all using a log-likelihood reward). We identified a bias in GRPO's advantage estimate due to group standard normalization as the relevant difference between these algorithms and provided a theoretical explanation for why normalization causes overconfidence. We also found that using the clipped policy gradient introduced by PPO did not impact calibration in our experiments.

Our results fit into a broader set of findings that biased policy gradients can lead to unexpected behavior for reasoning language models. For example, \citet{drgrpo} introduce Dr. GRPO, a modification of GRPO designed to eliminate terms that introduce bias. They propose to remove length normalization, which they find biases models to longer outputs, and to remove group standard normalization, which they interpret as a question-level difficulty bias. Our work identifies a novel negative impact of standard normalization in GRPO and supports unbiasedness as a useful design principle for policy gradient methods in reasoning RL.

We note that there are other possible framings of the outcome probability task explored in this paper. For example, one could directly estimate the probability of stochastic outcomes and train models to accurately predict these continuous values. While summarizing uncertainty can be useful, this approach requires having robust probability estimates ahead of time, which may be unavailable or model dependent, and limits the opportunity for the reasoning model to learn to make more precise estimates. Alternatively, one can train only on deterministic tasks and hope for transfer to stochastic settings (for example, we observe better than random zero-shot predictions on the CRISPR task in Appendix Fig.~\ref{fig:crispr_zeroshot}), though this limits the available data and tasks for training models. Overall, we believe that modeling stochastic outcomes from observed samples is an important capability for reasoning RL and that it is useful to characterize algorithms for this setting.

Finally, we presented an initial application of RL to train reasoning models directly from noisy biological experiments. While we found that RL can yield calibrated predictions for held-out experiments, these predictions do not necessarily reflect rigorous reasoning about uncertainty in the model's chain-of-thought. We are excited to further investigate how to train models to reason rigorously about uncertainty and support scientific discovery.

\section*{Acknowledgments}
Thanks to Neil Band, Marcel Rød, Shirley Wu, Rok Sosic, and Nick Haber for helpful comments and discussion.

\bibliography{colm2025_conference}
\bibliographystyle{colm2025_conference}

\newpage
\appendix
\section{Appendix}

\makeatletter
\renewcommand*{\theHfigure}{A\arabic{figure}}
\makeatother

\subsection{Analysis of Bias in GRPO Advantage Estimates} \label{bias_analysis}

Let $q$ be a prompt and $a \sim p(A|q) = \text{Bernoulli}(p)$ be a stochastic binary answer for the prompt. Let $\hat{p}$ be a predicted probability that $a=1$ from model $\pi_\theta$. Let $\mathrm{r}(\hat{p}, a)$ be a reward based on a strictly proper scoring rule such as the log-likelihood $\mathrm{r}(\hat{p}, a) = a\log\hat{p} + (1-a)\log (1-\hat{p})$ or Brier score $\mathrm{r}(\hat{p}, a) = -(a-\hat{p})^2$. Proper scoring rules have the property that the expected value is maximized by the true probability and have been shown to be effective rewards for training calibrated classifiers \citep{bandLinguisticCalibrationLongForm2024}.

The true advantage estimate for prompt $q$ and prediction $\hat{p}$ is
\begin{align*}
    A(q, \hat{p}) &= Q^{\pi}(q, \hat{p}) - V^{\pi}(q) \\
        &= \mathbb{E}_{a\sim p(A|q)}[\mathrm{r}(\hat{p}, a)] - \mathbb{E}_{a\sim p(A|q), \hat{p}' \sim \pi_\theta(q)}[\mathrm{r}(\hat{p}', a)] \\
        &= p \mathrm{r}(\hat{p}, 1) + (1-p)\mathrm{r}(\hat{p}, 0) - \mathbb{E}_{\hat{p}'} \left[ p\mathrm{r}(\hat{p}', 1) + (1-p)\mathrm{r}(\hat{p}', 0) \right]\\
        &= p (\mathrm{r}(\hat{p}, 1) + \mu_1) + (1-p)(\mathrm{r}(\hat{p}, 0) + \mu_0)
\end{align*}

where $\mu_1 = \mathbb{E}_{\hat{p}' \sim \pi_\theta(q)}[\mathrm{r}(\hat{p}', 1)]$ and $\mu_0 = \mathbb{E}_{\hat{p}' \sim \pi_\theta(q)}[\mathrm{r}(\hat{p}', 0)]$.

Next, we compare the advantage estimates from GRPO \citep{deepseekmath} to the true advantage to characterize any biases. Without loss of generality, we will set the index of the prediction whose advantage we are estimating to $i$ ($\hat{p} = \hat{p}_i$) and define $\hat{\mathbf{p}} = (\hat{p}_1, ..., \hat{p}_G) \sim \pi_\theta(q)$ to be the set of $G$ predictions in the group sampled from the same prompt $q$. We define $\hat{\mathbf{p}}_{i\neq j}$ to be the predictions in the group other than $\hat{p}_i$. We see that the expected advantage for GRPO without standard normalization is

\begin{align*}
    \mathbb{E}_{\substack{a \sim p(A|q), \\\hat{\mathbf{p}}_{j \neq i} \sim \pi_\theta(q)}} \left[ \hat{A}^{\text{NO-STD}}(q, \hat{p}) \right] &= \mathbb{E}_{a, \hat{\mathbf{p}}_{j \neq i}}\left[ \mathrm{r}(\hat{p}_i, a) - \text{mean}(\mathrm{r}(\hat{\mathbf{p}}, a)) \right] \\
        &= \mathbb{E}_{a, \hat{\mathbf{p}}_{j\neq i}}\left[\mathrm{r}(\hat{p}_i, a) - \frac{1}{G}\left(\mathrm{r}(\hat{p}_i, a) + \sum_{j \neq i} \mathrm{r}(\hat{p}_j, a)\right) \right]\\
        &= \mathbb{E}_{a} \left[ \left(\mathrm{r}(\hat{p}_i, a)  - \frac{1}{G} \mathrm{r}(\hat{p}_i, a)\right) - \frac{G-1}{G} \mathbb{E}_{\hat{p}'\sim \pi_\theta(q)} \mathrm{r}(\hat{p}', a)\right] \\
        &= \frac{G-1}{G} \mathbb{E}_a \left[ \mathrm{r}(\hat{p}_i, a) - \mathbb{E}_{\hat{p}'} [\mathrm{r}(\hat{p}', a)]\right]\\
        &= \frac{G-1}{G} A(q, \hat{p})
\end{align*}

We see that the estimate is proportional to the true advantage, though it is attenuated by a factor of $\frac{1}{G}$. A fully unbiased estimate can be achieved with the advantage from RLOO \citep{rloo1}, which exclude $\hat{p}_i$ from the mean baseline.

Finally, we consider the expected GRPO advantage estimate with standard normalization. We define $\sigma_1 = \mathbb{E}_{\hat{\mathbf{p}} \sim \pi_\theta(q)} \left[\text{std}(\mathrm{r}(\hat{\mathbf{p}}, 1))\right]$ and $\sigma_0 = \mathbb{E}_{\hat{\mathbf{p}} \sim \pi_\theta(q)} \left[\text{std}(\mathrm{r}(\hat{\mathbf{p}}, 0))\right]$. We make the simplifying assumption that group size $G$ is large so that $\mathbb{E}_{\hat{\mathbf{p}}_{j \neq i} \sim \pi_\theta(q)} \left[\text{std}(\mathrm{r}(\hat{\mathbf{p}}, 1))\right] \approx \sigma_1$ and $\mathbb{E}_{\hat{\mathbf{p}}_{j \neq i} \sim \pi_\theta(q)} \left[\text{std}(\mathrm{r}(\hat{\mathbf{p}}, 0))\right] \approx \sigma_0$, and will ignore the dependency between the mean and standard deviation of group rewards. We have:

\begin{align*}
   \mathbb{E}_{\substack{a \sim p(A|q), \\\hat{\mathbf{p}}_{j \neq i} \sim \pi_\theta(q)}} \left[ \hat{A}^{\text{STD}}(q, \hat{p}) \right] &= \mathbb{E}_{a, \hat{\mathbf{p}}_{j \neq i}} \left[\frac{\mathrm{r}(\hat{p}, a) - \text{mean}(\mathrm{r}(\hat{\mathbf{p}}, a))}{\text{std}(\mathrm{r}(\hat{\mathbf{p}}, a)) + \epsilon} \right] \\
    &\approx p \frac{\mathrm{r}(\hat{p}, 1)- \mu_1}{\sigma_1 + \epsilon} + (1-p) \frac{\mathrm{r}(\hat{p}, 0) - \mu_0}{\sigma_0 + \epsilon} \\
    &= \frac{1}{\sigma_1 + \epsilon} p (\mathrm{r}(\hat{p}, 1) + \mu_1) + \frac{1}{\sigma_0 + \epsilon} (1-p)(\mathrm{r}(\hat{p}, 0) + \mu_0)
\end{align*}

We see that the approximate expected advantage estimate from GRPO has the same weighted reward terms as the true advantage with the addition of new $\frac{1}{\sigma_1 + \epsilon}$ and $\frac{1}{\sigma_0 + \epsilon}$ coefficients. These coefficients make the GRPO advantage estimate biased in a policy-dependent manner, which is analyzed in Figures~\ref{fig:cause}, \ref{fig:brier_advantage_estimates}, and \ref{fig:explanation_sigma}.

\subsection{Advantage Empirical Estimate Details} \label{empirical_advantage_details}

We compute the empirical GRPO advantage estimates in Fig.~\ref{fig:cause} for a log-likelihood reward using group size $G=1000$, true probability $p=0.7$, and 100,000 samples of $\mathbf{\hat{p}} = (\hat{p}_1, ..., \hat{p}_G) \sim \pi_\theta$. Policies are categorical distributions over predicted probabilities $(0.01, 0.02, ..., 0.99)$, where categorical log probs are set by discretizing Beta distributions ($\text{Beta}(1,1)$, $\text{Beta}(5.7,3)$, $\text{Beta}(50, 1)$). Empirical advantage estimates are plotted for predictions with at least 1,000 observed samples. True advantages are computed exactly. 

\subsection{GRPO Bias with Other Rewards}
\label{other_reward_bias}

\begin{figure}[h]
    \centering
    \includegraphics[width=0.8\linewidth]{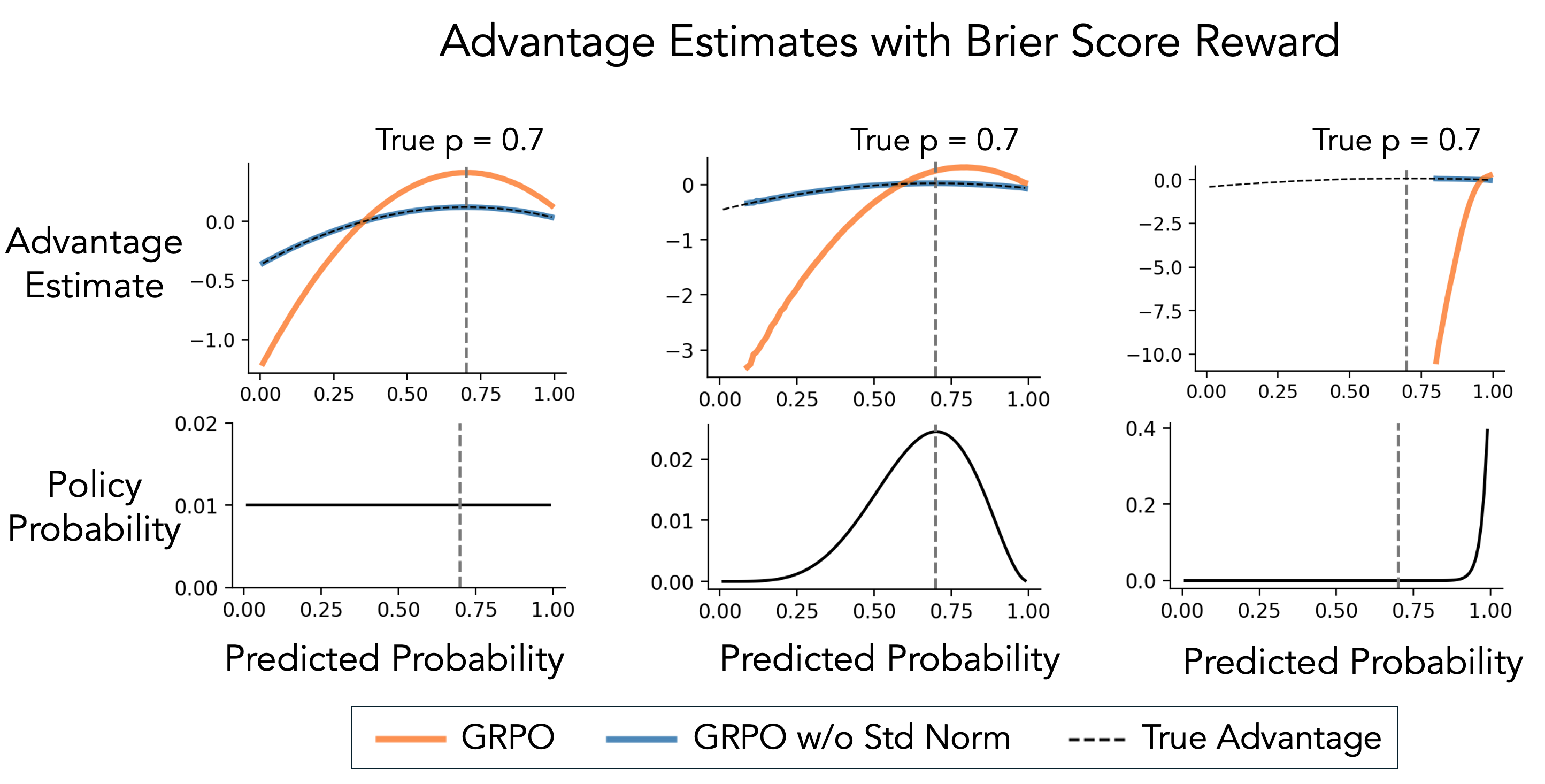}
    \caption{Analysis of advantage estimates with a reward based on the Brier score. We observe a similar pattern of overestimated advantages for overconfident probabilities as observed with a log-likelihood in Fig.~\ref{fig:cause}}.
    \label{fig:brier_advantage_estimates}
\end{figure}

\begin{figure}[h]
    \centering
    \includegraphics[width=0.8\linewidth]{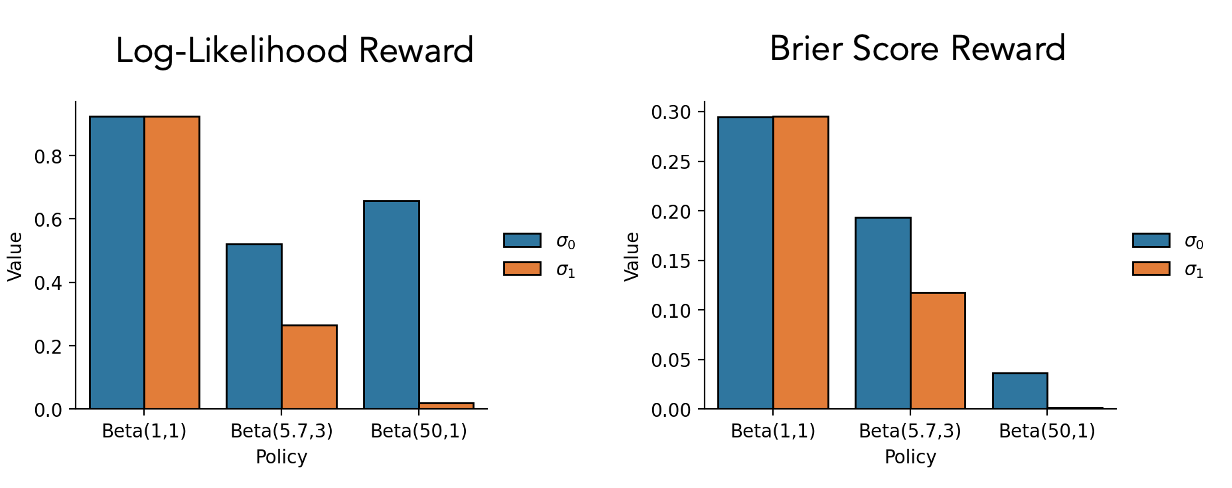}
    \caption{Empirical estimates of $\sigma_0$ and $\sigma_1$ (standard deviation of rewards within groups for answers 0 and 1) for the three policies in Figures \ref{fig:cause} and \ref{fig:brier_advantage_estimates}. As the policies concentrate on predictions greater than 0.5, $\sigma_0$ becomes larger than $\sigma_1$.}
    \label{fig:explanation_sigma}
\end{figure}

While our analysis in the main text focused on a log-likelihood reward, prior work has found that optimization based on other proper scoring rules (which are maximized in expectation by the true probability) can yield well-calibrated classifiers \citep{bandLinguisticCalibrationLongForm2024}. We show a similar pattern of GRPO advantage estimate biases with a reward based on the Brier score $\mathrm{r(\hat{p}, a)} = -(a-\hat{p})^2$ in Fig.~\ref{fig:brier_advantage_estimates}. We hypothesize that GRPO will yield overconfident predictions for rewards based on strictly proper scoring rules more generally because they are strictly concave, which should lead to similar changes in $\sigma_1$ and $\sigma_2$ as the policy changes, but we leave a formal characterization as out of scope for this paper.

\FloatBarrier

\subsection{Synthetic Data Experiment Extended Results}

\begin{table}[h]
\centering

\begin{tabular}{lccccc}
    \toprule
    Algorithm & Grad Steps / Rollout & $\epsilon_{\text{clip}}$ & ECE & AUROC & Accuracy \\
    \midrule
    GRPO & 1 & NA & 0.239 & 0.750 & 0.751 \\
    GRPO & 10 & 0.200 & 0.239 & 0.751 & 0.751 \\
    GRPO & 10 & 0.001 & 0.239 & 0.751 & 0.751 \\
    \midrule
    GRPO (No Std) & 1 & NA & 0.002 & 0.823 & 0.751 \\
    GRPO (No Std) & 10 & 0.200 & 0.005 & 0.823 & 0.751 \\
    GRPO (No Std) & 10 & 0.001 & 0.005 & 0.823 & 0.751 \\
    \midrule
    PPO & 1 & NA & 0.005 & 0.823 & 0.751 \\
    PPO & 10 & 0.200 & 0.008 & 0.823 & 0.751 \\
    PPO & 10 & 0.001 & 0.008 & 0.823 & 0.751 \\
    \midrule
    RLOO & 1 & NA & 0.002 & 0.823 & 0.751 \\
    RLOO & 10 & 0.200 & 0.004 & 0.823 & 0.751 \\
    RLOO & 10 & 0.001 & 0.004 & 0.823 & 0.751 \\
    \bottomrule
    \end{tabular}
    \caption{Extended results from synthetic data experiments. We observe that results are consistent between experiments with a single update per rollout and multiple updates per rollout with a clipped policy gradient estimates, even with low clipping thresholds that encourage high clipping rates.}
    \label{tab:extended_synthetic_results}
\end{table}

\begin{figure}[h]
    \centering
    \includegraphics[width=\linewidth]{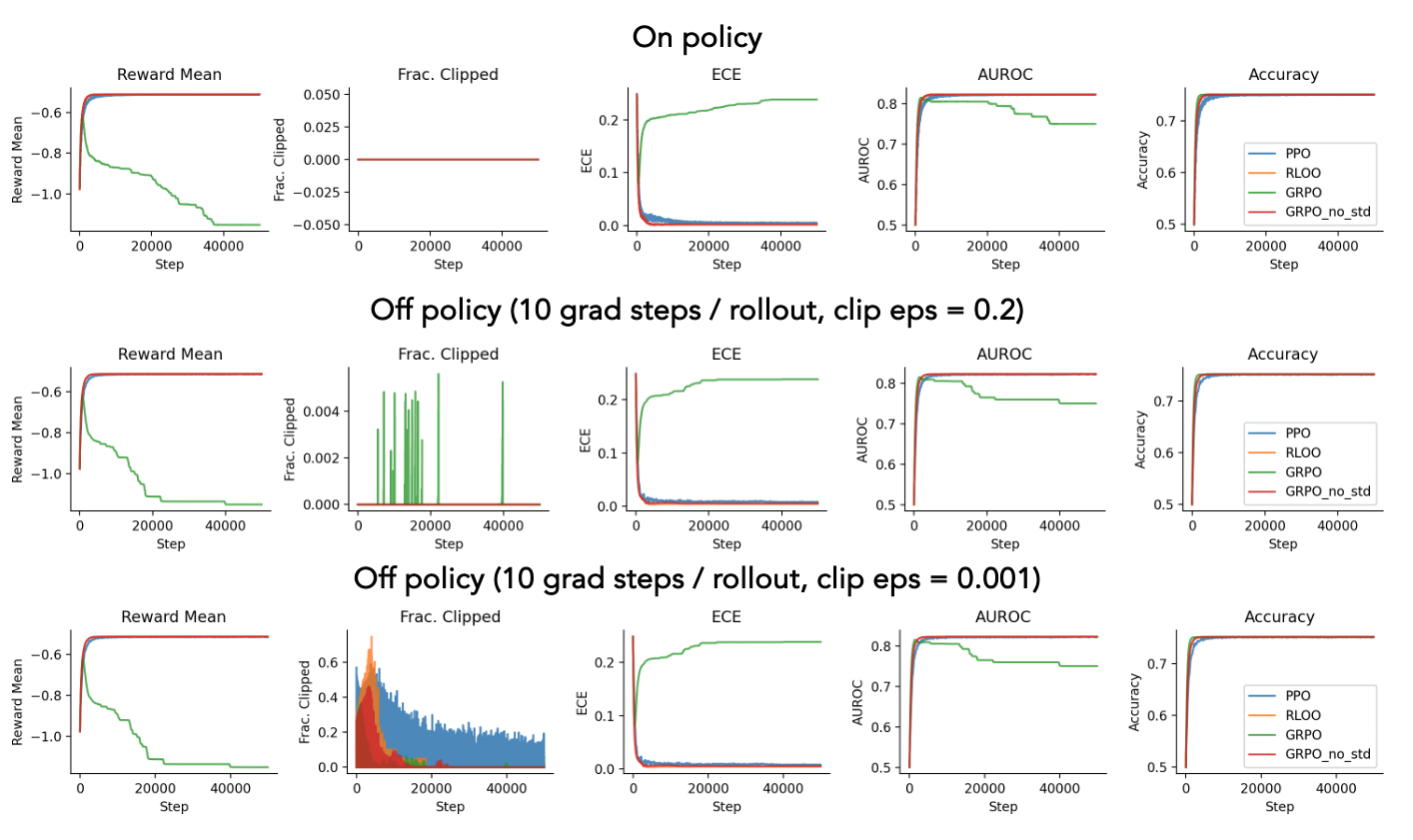}
    \caption{Synthetic data training metrics.}
    \label{fig:synthetic_training_metrics}
\end{figure}

\newpage
\subsection{CRISPR Experiment Data Processing}
\label{crispr_preprocessing}

CRISPR perturb-seq screens involve perturbing individual genes with CRISPR (which modulates the expression of a target gene) and measuring the effect of the perturbation on RNA transcript counts for all genes in individual cells cell. We use the essential gene CRISPRi (CRISPR interference) perturb-seq screen in K562 cells from \citet{replogle} for our experiment. The dataset contains CRISPRi perturbations, which lower gene expression, that target approximately 2,000 unique genes. We apply consensus non-negative matrix factorization (cNMF) \citep{cnmf} to infer 50 aggregate transcriptional target phenotypes and select the top 15 marker genes for each phenotype as defined by the cNMF method to describe each phenotype. We estimate the effect size of each perturbation on each phenotype as the difference in mean phenotype values for cells that received the perturbation and control cells. To define perturbations with strong effects ("hits"), we fit a cluster model on the perturbation effect sizes for each phenotype and select perturbations that are highly unlikely in the control cluster. Specifically, we fit a Gaussian Mixture Model on the effect sizes for each phenotype (number of clusters between 1-4, selected based on Bayesian Information Criterion) and select perturbations with $<$1\% chance under the cluster closest to zero as strong effects. To construct a balanced dataset, we select an equal number of perturbations that are most likely under the control cluster as non-hits. We note that the dataset is naturally very imbalanced (hits are relatively rare for most phenotypes) but choose to work with a balanced dataset for simplicity as our primary focus is understanding the behavior of RL algorithms with stochastic outcomes.

\subsection{CRISPR Task Prompt}
\label{crispr_prompt}

\begin{tcolorbox}[
    enhanced,
    breakable,
    colback=blue!3,
    colframe=blue!50!black,
    title=Experiment Prediction Prompt,
]
\begin{Verbatim}[breaklines, breakanywhere, fontsize=\small]
I am planning a perturb-seq screen and plan to assess effects of perturbations on a phenotype with the following marker genes: ```{pheno_markers}```.

How likely is a CRISPRi perturbation applied to {pert} to have a strong effect on this phenotype? Respond with probability from 1-99, representing 1% to 99% chance of a strong effect. Enclose your answer in <answer> </answer> tags.
\end{Verbatim}
\end{tcolorbox}

\texttt{pheno\_markers} is a list of 15 marker genes for the phenotype, and \texttt{pert} is the gene perturbed by the CRISPR perturbation. We also considered prompts that specified the overall frequency of hits in the dataset, but found that this reduced the zero-shot model performance.

\subsection{CRISPR Experiment Details}
\label{crispr_experiment_details}

\begin{figure}[h]
    \centering
    \includegraphics[width=1\linewidth]{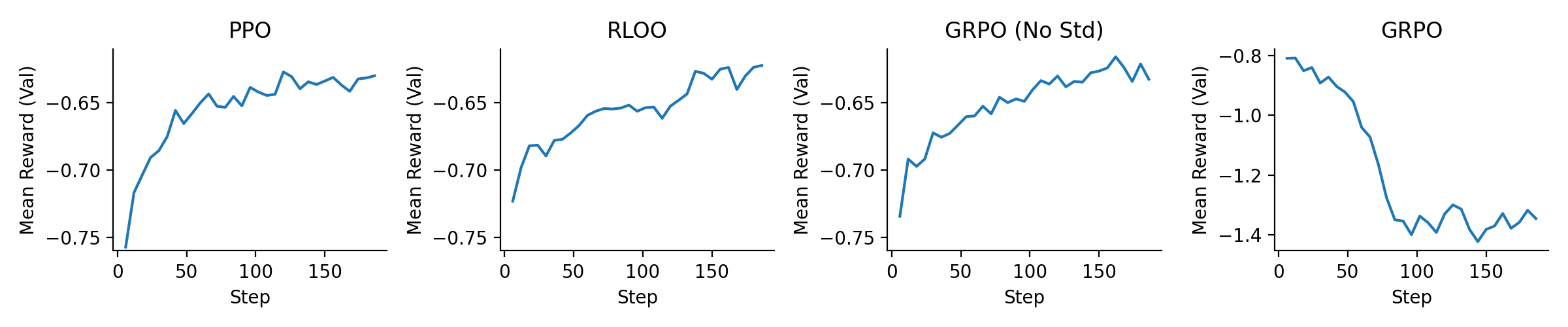}
    \caption{CRISPR experiment prediction task validation set rewards during training.}
    \label{fig:pert_val_rewards}
\end{figure}

Models were trained with a log-likelihood reward, with a minimum reward of $\log 0.01$ for outputs that do not match the required format (corresponding to the worst possible reward given the prediction range of 0.01-0.99). Each model was trained with batch size 512, group size 4, max response length 2048, mini-batch size (batches for gradient updates within each rollout) of 64, learning rate 1e-6, and KL loss coefficient of 0.001. For PPO, the critic is trained with mini-batch size 64 and learning rate 1e-5. We train all models without length normalization as discussed in \citet{drgrpo} to avoid a length bias. Models were trained for 16 epochs with Verl \citep{verl}. For PPO, RLOO, and GRPO with no standard normalization, we select the checkpoint with the best validation reward for evaluation (epoch 15 / step 180 for all three). We use the same checkpoing from the GRPO run for consistency (validation reward begins dropping early and we want to understand what predictions it converges to) (Fig.~\ref{fig:pert_val_rewards}). We generate 4 samples per prompt for test set evaluation and drop samples with no valid prediction (at most one sample of 5608 predictions for each trained models).

\begin{figure}[h]
    \centering
    \includegraphics[width=0.35\linewidth]{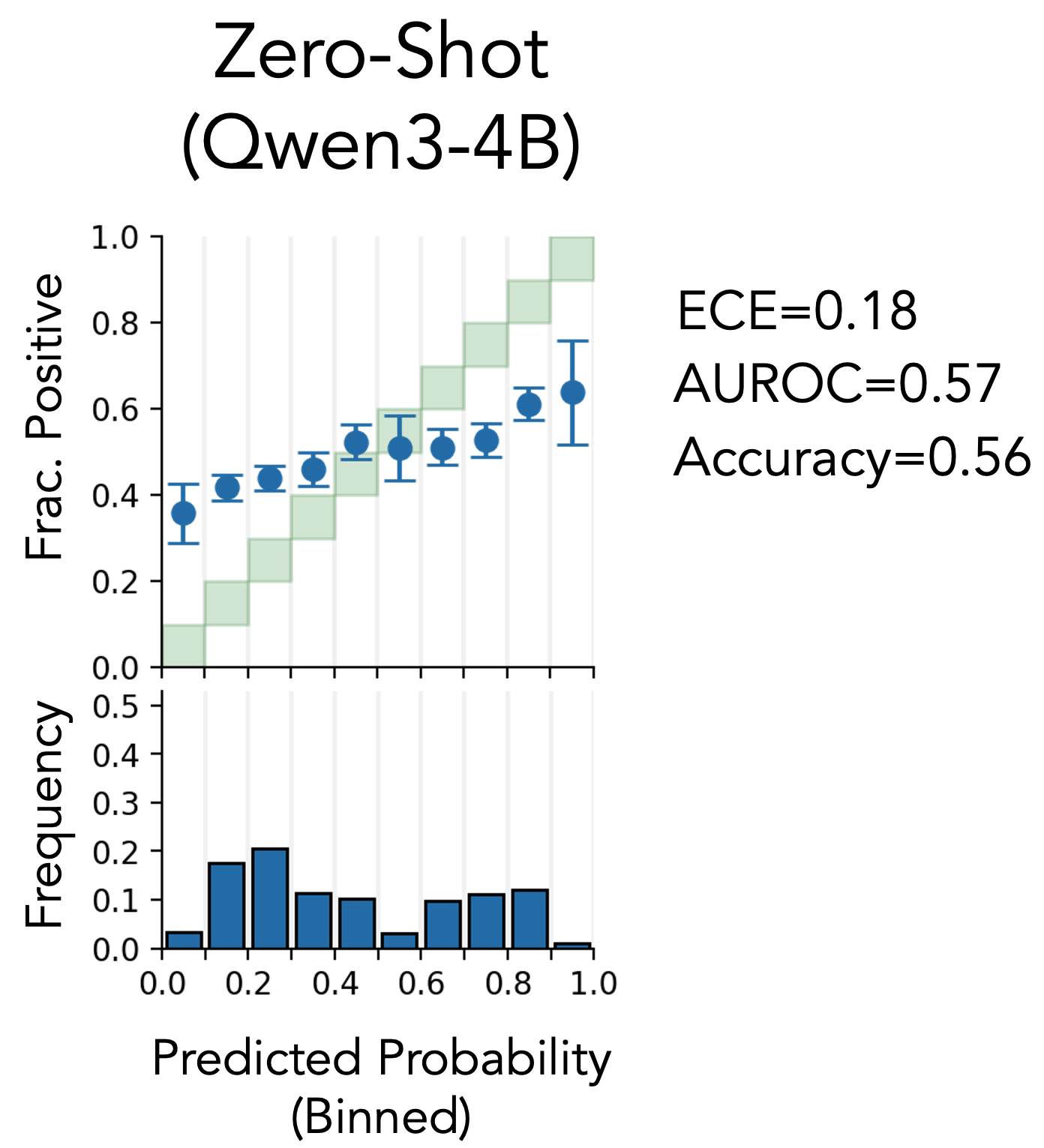}
    \caption{Zero shot predictions on CRISPR task test set with Qwen3-4B.}
    \label{fig:crispr_zeroshot}
\end{figure}

\end{document}